\documentclass[conference]{IEEEtran}
\usepackage{siunitx}
\usepackage{booktabs}
\usepackage{comment}
\usepackage{multirow}
\usepackage{numprint}
\usepackage{cite}
\usepackage{amsmath,amssymb,amsfonts}
\usepackage{algorithmic}
\usepackage{graphicx}
\usepackage{textcomp}
\usepackage{xcolor}
\usepackage[nolist]{acronym}
\usepackage[hidelinks]{hyperref}

\def\BibTeX{{\rm B\kern-.05em{\sc i\kern-.025em b}\kern-.08em
    T\kern-.1667em\lower.7ex\hbox{E}\kern-.125emX}}
\begin{document}
\newacro{ml}[ML]{machine learning}
\newacro{fd}[FD]{fault detection}
\newacro{fli}[FLI]{fault line identification}
\newacro{pr}[PR]{protection relay}
\newacro{pc}[PC]{protection coordination}
\newacro{rf}[RF]{Random Forest}

\title{Impact of Data Sparsity on Machine Learning for Fault Detection in Power System Protection}

\author{
\IEEEauthorblockN{Julian Oelhaf\textsuperscript{1}\textsuperscript{*}, Georg Kordowich\textsuperscript{2}, Changhun Kim\textsuperscript{1}, Paula Andrea Pérez-Toro\textsuperscript{1}, Andreas Maier\textsuperscript{1} \\
Johann Jäger\textsuperscript{2}, Siming Bayer\textsuperscript{1}}
\IEEEauthorblockA{\textit{\textsuperscript{1}Pattern Recognition Lab, Friedrich-Alexander-University Erlangen-Nuremberg} \\
\textit{\textsuperscript{2}Institute of Electrical Energy Systems, Friedrich-Alexander-University Erlangen-Nuremberg} \\
Erlangen, Germany \\
\textsuperscript{*}corresponding author: julian.oelhaf@fau.de}
}

\maketitle

\begin{abstract}
Germany's transition to a renewable energy-based power system is reshaping grid operations, requiring advanced monitoring and control to manage decentralized generation. Machine learning (ML) has emerged as a powerful tool for power system protection, particularly for fault detection (FD) and fault line identification (FLI) in transmission grids. However, ML model reliability depends on data quality and availability.
Data sparsity resulting from sensor failures, communication disruptions, or reduced sampling rates poses a challenge to ML-based FD and FLI. Yet, its impact has not been systematically validated prior to this work.
In response, we propose a framework to assess the impact of data sparsity on ML-based FD and FLI performance.
We simulate realistic data sparsity scenarios, evaluate their impact, derive quantitative insights, and demonstrate the effectiveness of this evaluation strategy by applying it to an existing ML-based framework.  
After 50-fold decimation, the model achieves an F1-score of 0.999 $\pm$ 0.000 for FD. For FLI, removing current measurements reduces the F1-score by 55.61\%, and the Bus~2 loss scenario reduces it by 9.73\%. These results quantify the sensitivity of the two tasks to the evaluated input restrictions.
\end{abstract}


\section{Introduction}

% Context: Germany's energy transition (Energiewende) and its impact on grid operations.  
Germany's \textit{Energiewende}, the transition to a renewable energy-based power system, is fundamentally reshaping electric grid operations. The growth of wind, solar, and hydroelectric generation, along with the increase in energy efficiency and demand-side management, presents new challenges for ensuring a secure, stable, and affordable electricity supply~\cite{kabeyi_sustainable_2022}. A key aspect of this transition is the expansion and modernization of the power transmission infrastructure, spanning thousands of kilometers, to enable the integration of fluctuating renewable energy while ensuring robust grid operations~\cite{bmwk2024}.

% Challenge: Decentralized energy requires new control and monitoring approaches.  
As the power system undergoes this transformation, the grid itself must evolve to accommodate \textit{decentralized} energy sources, leading to \textit{dynamic and bidirectional} power flows. Traditional grid control mechanisms, designed for centralized generation, are increasingly insufficient to manage the variability and distributed nature of renewable energy sources~\cite{central_power_research_institute_protection_2017}. To address these challenges, digitalization has become an essential component of modern grid infrastructure, driving an unprecedented proliferation of data from a vast network of sensors and measurement devices. These data streams originate from \acp{pr} and other monitoring systems, which transfer real-time information to centralized control centers for analysis and decision-making~\cite{aftab_iec_2020}.
A key enabler of this transformation is the \textit{IEC 61850} communication standard~\cite{international_electrotechnical_commission_iec_2024}, which enables the integration of remote measurements from protection systems. By improving fault monitoring and enhancing grid resilience, \textit{IEC 61850} plays a crucial role in ensuring efficient and reliable power system operation~\cite{suhail_hussain_novel_2016}. However, as the volume and complexity of grid data continue to grow, traditional analytical approaches struggle to extract actionable insights in real-time.

% Solution: ML improves fault detection but relies on high-quality data.  
To address this challenge, \ac{ml} models have emerged as a promising solution for \ac{fd}~\cite{suhail_hussain_novel_2016, abdullah_ultrafast_2018, oelhaf25}, \ac{fli}~\cite{zainab_faulted_2019}, and \ac{pc}~\cite{kordowich_hybrid_2022}, enabling more accurate and adaptive fault management while efficiently handling large-scale grid data. However, the performance of \ac{ml} methods is highly dependent on the quality and availability of training data. An over-reliance on large datasets can introduce vulnerabilities, particularly when these datasets do not adequately reflect real-world data sparsity caused by sensor failures, communication interruptions, and hardware limitations. Such sparsity can compromise \ac{ml}-based \ac{fd} and \ac{fli} systems, potentially leading to delayed or incorrect protection actions.

% Key Factors Affecting Data Availability
To systematically assess the impact of data sparsity, it is essential to consider the factors influencing data availability in power system protection, such as sampling frequency and feature availability. Higher sampling frequencies provide more precise waveform information, but real-world devices often operate at lower rates. Similarly, feature availability, which depends on sensor placement and grid topology, can lead to missing or degraded measurements. In practice, sampling rates typically range from \(\qty{1}{\kilo\hertz}\) to \(\qty{10}{\kilo\hertz}\), varying by relay type and manufacturer. Additionally, legacy infrastructure and communication delays often complicate data availability.

% Simulations as a Necessary but Imperfect Tool
% Gap Between Research Assumptions and Real-World Constraints
Due to the challenges in obtaining real-world high-resolution fault data, research has relied on simulations. Simulations offer the ability to generate extensive datasets and replicate rare fault scenarios.
Many studies~\cite{chingshom_fault_2024,jabbar_power_2024,najafzadeh_fault_2024} on \ac{fd} and \ac{fli} simulate data sampled at \(\qty{20}{\kilo\hertz}\), which can often be unrealistic in practice. However, the applicability of such studies depends on whether the simulated data truly reflects real-world grid behavior. Addressing the limitations of current \ac{ml}-based \ac{fd} and \ac{fli} models under data-sparse conditions is crucial for their real-world deployment. A critical gap remains: there is no systematic framework to validate the performance of ML-based protection schemes under varying data quality and availability constraints.

% Define the Research Gap and Contribution 
To the best of our knowledge, no prior work has introduced a validation framework to assess the impact of data sparsity on \ac{ml}-based \ac{fd}.
This study addresses this gap by presenting a proof of concept for evaluating \ac{ml} models under real-world data constraints with an existing \ac{ml} model.
We systematically analyze how missing sensor data, reduced sampling rates, and temporary communication failures affect \ac{ml} performance.
Focusing on three-phase short circuits within a single power grid topology, we demonstrate the feasibility of a structured validation approach for data quality and availability.
Our findings lay the groundwork for assessing the robustness of ML-based \ac{fd} and \ac{fli} methods, providing insights to improve their reliability and real-world deployment in data-limited scenarios.

\section{Methods and Materials} \label{sec:methodology}

\subsection{Dataset and Grid Topology}\label{sec:dataset}
\begin{figure*}[!htbp]
	\centering
	\includegraphics[trim={0 0 0 0},clip,width=0.98\linewidth]{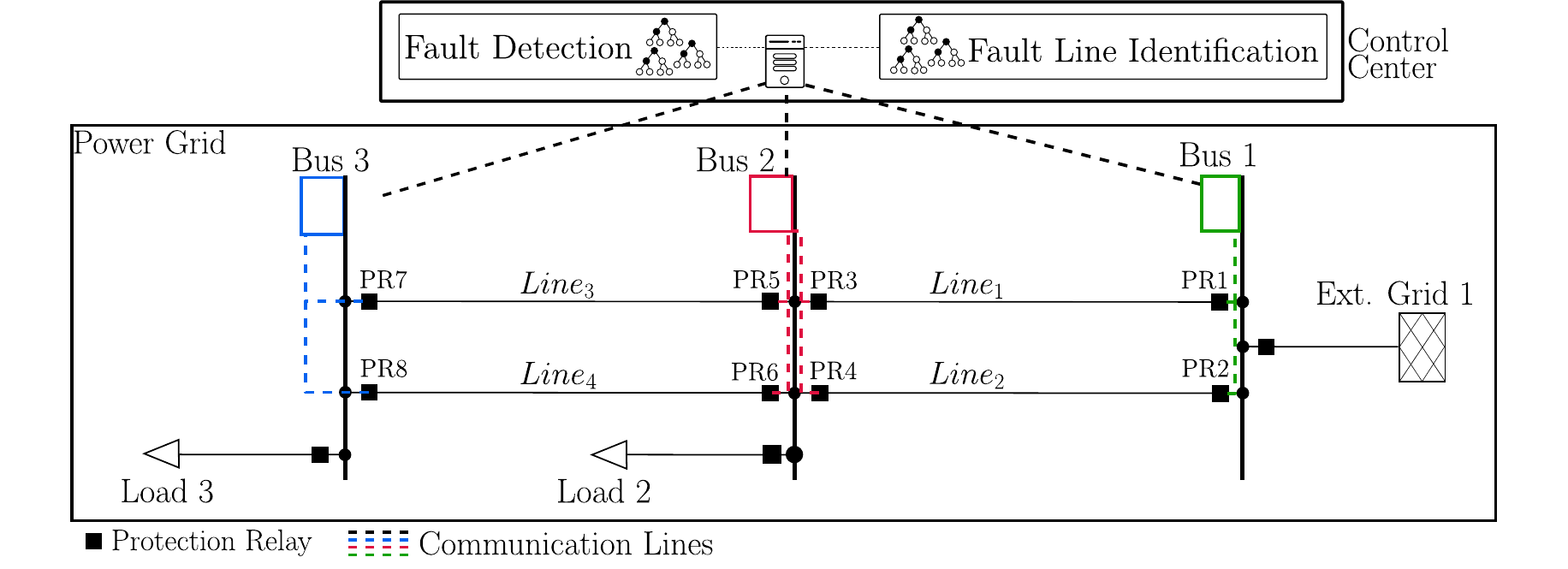}
    \caption{Schematic of the ``Double Line'' grid topology. The diagram illustrates the transmission lines, \acp{pr}, and communication devices. Measurements are recorded at the \acp{pr} and transmitted to the substations at each bus. From there, the data is relayed to the control center for \ac{fd} and \ac{fli}.}
    \label{fig:double_line_grid_topology}
    % \vspace{-2mm} % Adjust vertical spacing if necessary
\end{figure*}

To investigate the impact of data sparsity, we use a widely adopted topology in grid protection studies, as outlined in~\cite{meyer_hybrid_2020}. The power grid simulation is based on the ``Double Line'' topology, shown in Fig.~\ref{fig:double_line_grid_topology}, and generated using \mbox{DIgSILENT's} PowerFactory\footnote{\href{https://www.digsilent.de/en/}{https://www.digsilent.de/en/}} software, with simulations conducted by an expert in electrical power engineering. This framework extends the methodology from~\cite{wang_generic_2022}.
Key parameters, including external grid settings, load conditions, transmission line lengths, and short circuit locations, were systematically varied to simulate a broad range of fault scenarios. The rationale behind this parameter variation, known as domain randomization, is to improve the robustness of \ac{ml} training and facilitate the transition from simulation to real-world scenarios. The selected parameters aim to accurately represent realistic grid conditions while covering a wide range of possible scenarios. The parameter ranges were based on the works by~\cite{roeper_kurzschlusstrome_1984} and~\cite{oeding_elektrische_2016}. For further details on the parameter randomization process, refer to our previous study~\cite{oelhaf25}.
This study specifically focuses on three-phase faults, which, although accounting for only \(5\%\) of fault occurrences~\cite{gonen_electric_2015}, are responsible for the most severe disruptions in the grid.
The objective is to assess whether these faults can be reliably detected despite the challenges posed by data sparsity.

Each sample has a unique fault configuration, lasting \(\qty{1}{\second}\) at a \(\qty{20}{\kilo\hertz}\) sampling rate. Since \acp{pr} installed at Buses measure three-phase current and voltage \(\left(I_A, I_B, I_C, V_A, V_B, V_C\right)\), each device records a total of six signals. Given that eight \acp{pr} are deployed on the transmission lines, the system generates \(\qty{48}{} \) measurements per time step, serving as input for \ac{fd} and \ac{fli}.

In total, the network consists of three Buses (Bus 1–3) connected by four transmission lines. Bus 1 links to an external grid, while Bus 2 acts as a junction with four \acp{pr} monitoring two incoming and two outgoing lines. Bus 3 is connected to two transmission lines with \acp{pr}. In this study, we focus on the protection of transmission lines.

Current and voltage signals at each \ac{pr} are continuously monitored for real-time protection and \ac{fd}. These measurements form a multivariate time series essential for \ac{fd} and \ac{fli}, with the current and voltage signals for each phase given by:
\begin{align}
    I_{PR}(t) & = (I_A(t), I_B(t), I_C(t)),& t &\in [0, 1]\, \text{s}  \label{eq:IPR} \\ 
    V_{PR}(t) & = (V_A(t), V_B(t), V_C(t)), & t &\in [0, 1]\, \text{s} \label{eq:VPR}
\end{align}

where subscripts \(A, B, C\) correspond to the three phases. The combined multivariate time series per \ac{pr} is then:

\begin{align}
    X_{PR}(t) &=
    \begin{bmatrix}
    I_{PR}(t) \\
    V_{PR}(t)
    \end{bmatrix}, & t &\in [0, 1]\, \text{s} \label{eq:X_PR}
\end{align}

% Data Preprocessing
\subsection{Data Preprocessing} %% Preprocessing to simulate Real-Time Setting

We preprocess the raw simulation data by trimming each sample to \(\pm \qty{80}{\milli\second}\) around the fault start to capture critical events before and after the fault. To simulate real-time constraints, we slide the time series with a \(\qty{5}{\milli\second}\) step, ensuring overlap between snippets. We evaluate the impact of different window lengths: \(\qty{10}{\milli\second}\), \(\qty{20}{\milli\second}\), \(\qty{30}{\milli\second}\), \(\qty{40}{\milli\second}\), and \(\qty{50}{\milli\second}\).
These window lengths were chosen to balance between capturing enough information for accurate \ac{fd} while adhering to the practical constraints of real-time processing. Additionally, with a \(\qty{50}{\hertz}\) power system, one period corresponds to a duration of \(\qty{20}{\milli\second}\).

\subsection{Machine Learning Models for Fault Detection and Line Identification}

\ac{fd} is formulated as a binary classification task, while \ac{fli} as a four-class classification problem. The feature space consists of concatenated voltage and current measurements from the four transmission lines, forming a multivariate time series over a predefined window length. Each window is labeled based on the presence or absence of a three-phase short-circuit fault, and, where applicable, the corresponding faulted transmission line, as described in~\cite{oelhaf25}. 

For consistency across all tasks, the same \ac{ml} model is used in all experiments. The \ac{rf} classifier is selected based on our comparative study~\cite{oelhaf25}, which demonstrated its favorable trade-off between performance and runtime. The reported F1-score is \(0.998\) for \ac{fd}, \(0.985\) for \ac{fli}, with a runtime of \(\qty{16.8}{\milli\second}\).

The evaluation uses \numprint{4000} simulated episodes generated using the procedure outlined in Sec.~\ref{sec:dataset}. All experiments follow a 5-fold cross-validation scheme, ensuring consistent evaluation on unseen data. The F1-score is used as the evaluation metric due to its balanced consideration of both sensitivity and specificity. Model implementation, training, cross-validation, and evaluation are carried out using \textit{scikit-learn}~\cite{scikit_learn_2015}.

\subsection{Scenarios of Data Sparsity}  
\label{sec:data_sparsity}  

This study examines the impact of data sparsity on \ac{fd} and \ac{fli} in power systems by simulating realistic cases of missing or incomplete data. In collaboration with an expert in electrical power engineering, we identified plausible scenarios that may occur in a power grid, including hardware failures, sensor malfunctions, and communication disruptions. Additionally, we account for legacy infrastructure operating at lower sampling rates. Each scenario is described in detail below.  

\subsubsection*{Missing Voltage or Current Data}
This scenario simulates sensor failures by setting either voltage or current measurements to zero while keeping the remaining data unchanged. Such failures can result from sensor malfunctions or communication issues.  

\subsubsection*{Reduced Temporal Resolution}
To evaluate reduced input sampling, we retain samples at regular intervals without an anti-alias filter. Starting from \(\qty{20}{\kilo\hertz}\), we decimate to \(\qty{10}{\kilo\hertz}\), \(\qty{4}{\kilo\hertz}\), \(\qty{2}{\kilo\hertz}\), \(\qty{800}{\hertz}\), and \(\qty{400}{\hertz}\). The scenario represents a reduction in retained or transmitted samples.

\subsubsection*{Component Failures}
This scenario is intended to model the failure of critical system components by removing all measurements from the affected component. The loss of communication with a substation would result in the loss of measurements from all \acp{pr} connected to the corresponding Bus. We model the failure for Buses 1, 2, and 3. Furthermore, we model the individual failure of each of the eight \acp{pr}. Buses and \acp{pr} are shown in Fig.~\ref{fig:double_line_grid_topology}.

\subsubsection*{Phase-Specific Data Loss}
This scenario is designed to mimic the loss of measurements from one of the phases in a three-phase system. Such a loss can be attributed to physical failures, such as damaged sensors or transmission lines, or to disruptions in communication. In this case, one of the three phases, Phase A, Phase B, or Phase C, experiences an outage, resulting in missing measurements.

\subsubsection*{Communication Loss}
We simulate temporary communication loss by setting measurements to zero over contiguous gaps from \(\qty{5}{\milli\second}\) to \(\qty{45}{\milli\second}\), in steps of \(\qty{5}{\milli\second}\). Each gap starts at \(10\%\) of the input window duration. We consider only gaps shorter than the window.

\begin{table}[htbp]
\centering
\caption{Overview of data sparsity scenarios and parameter variations.}
\resizebox{0.96\linewidth}{!}{
\begin{tabular}{|c|c|}
\hline
\textbf{Sparsity Scenario} & \textbf{Parameter Values} \\
\hline
Missing Voltage  & True, False \\
Missing Current  & True, False \\
\multirow{2}{*}{Reduced Sampling Frequency} & \(\qty{10}{\kilo\hertz}\), \(\qty{4}{\kilo\hertz}\), \(\qty{2}{\kilo\hertz}\) \\
 & \(\qty{800}{\hertz}\), \(\qty{400}{\hertz}\) \\
Substation Communication Failure  & 1, 2, 3 \\
Relay Communication Failure  & 1 - 8 \\
Phase Measurement Failure  & A, B, C \\
Temporal Communication Loss  & \(\qty{5}{\milli\second}\) - \(\qty{45}{\milli\second}\) \\
\hline
\end{tabular}}
\label{tab:param_var_overview}
\end{table}

Tab.~\ref{tab:param_var_overview} summarizes the different data sparsity scenarios and their corresponding parameter variations. Each scenario introduces a unique challenge for \ac{fd} models, helping us evaluate their robustness under real-world conditions.
By incorporating these variations into our study, we systematically assess how different types of data sparsity can affect \ac{fd} performance. This evaluation is crucial for understanding the resilience of \ac{ml} models in real-world applications, where missing or degraded data is a common challenge.

\section{Experiments and Results}

In this study, we perform experiments on two tasks: \ac{fd} and \ac{fli}. For each task, we evaluate five different window lengths and five data sparsity scenarios, as introduced in Sec.~\ref{sec:data_sparsity}, with the parameter values summarized in Tab.~\ref{tab:param_var_overview}. This results in a total of 270 experiment combinations. 

The complete-input reference F1-scores are \(0.9998 \pm 0.0001\) for \ac{fd} and \(0.997 \pm 0.001\) for \ac{fli}. For the reference and Tables~\ref{tab:missing_v_a_measurements} and~\ref{tab:bus_failure_results}, each score is the mean of the five window-specific cross-validation means, and the spread is the population standard deviation across those five means. We compute the relative change within each window length against its complete-input reference, then average the five changes at full precision and round for display. The tasks use different window populations and targets: \ac{fli} is multiclass and uses only fault windows, while \ac{fd} is binary.

\subsubsection*{Missing Voltage or Current Data}
Table~\ref{tab:missing_v_a_measurements} shows that the \ac{fd} F1-score changes by \(+0.02\%\) when current measurements are missing and by \(-0.16\%\) when voltage measurements are missing. For \ac{fli}, removing current measurements reduces the F1-score by \(55.61\%\), whereas removing voltage measurements changes it by \(+0.04\%\).

\begin{table}[htbp]
\centering
\caption{Impact of loss of voltage and current measurements on model performance. Changes are relative to complete inputs.}
\resizebox{0.98\linewidth}{!}{
\begin{tabular}{|c|c|c|c|}
\hline
\textbf{Task} & \textbf{Measurement} & \textbf{F1$\uparrow$} & \textbf{Change (\%)} \\
\hline
\multirow{2}{*}{\ac{fd}} & Missing \(I\) & \(1.000 \pm 0.000\) & \(+0.02\%\) \\
 & Missing \(V\) & \(0.998 \pm 0.001\) & \(-0.16\%\) \\
\hline
\multirow{2}{*}{\ac{fli}} &  Missing \(I\) & \underline{\(0.443 \pm 0.008\)} & \underline{\(-55.61\%\)} \\
 & Missing \(V\) & \(0.998 \pm 0.001\) & \(0.04\%\) \\
\hline
\end{tabular}}
\label{tab:missing_v_a_measurements}
\end{table}

\subsubsection*{Reduced Temporal Resolution}
For both \ac{fd} and \ac{fli}, the F1-score remained nearly unchanged even after 50-fold decimation to \(\qty{400}{\hertz}\).
The comparison measures performance with decimated inputs. Acquisition directly at a lower sampling rate was not evaluated.
The F1-score showed negligible improvements between \(0.01\%\) and \(0.09\%\). The largest change occurred in the \ac{fli} task at \(\qty{400}{\hertz}\), reaching an F1-score of \(0.998\) (\(+0.09\%\)). Due to space constraints, detailed results are omitted for this scenario.  

\subsubsection*{Component Failures}
In the bus-loss scenarios, \ac{fd} reaches a displayed F1-score of \(1.000\) for all three conditions. The \ac{fli} F1-score decreases by \(0.28\%\), \(9.73\%\), and \(2.84\%\) in the Bus~1, Bus~2, and Bus~3 conditions, respectively (Table~\ref{tab:bus_failure_results}).

\begin{table}[htbp]
\centering
\caption{Impact of Bus failures on model performance. Changes are relative to complete inputs.}
\resizebox{0.85\linewidth}{!}{
\begin{tabular}{|c|c|c|c|}
\hline
\textbf{Task} &\textbf{Bus Nr.} & \textbf{F1$\uparrow$} & \textbf{Change (\%)} \\
\hline
\multirow{3}{*}{\ac{fd}} & Bus 1 & \(1.000 \pm 0.000\) & \(+0.01\%\) \\
& Bus 2 & \(1.000 \pm 0.000\) & \(0.00\%\) \\
& Bus 3 & \(1.000 \pm 0.000\) & \(0.00\%\) \\
\hline
\multirow{3}{*}{\ac{fli}} &Bus 1 & \(0.995 \pm 0.001\) & \(-0.28\%\) \\
& Bus 2 & \underline{\(0.900 \pm 0.003\)} & \underline{\(-9.73\%\)} \\
& Bus 3 & \(0.969 \pm 0.002\) & \(-2.84\%\) \\
\hline
\end{tabular}}
\label{tab:bus_failure_results}
\end{table}

\subsubsection*{Phase-Specific Data Loss}
The absence of phase measurements had a negligible effect on the model's performance, with only minor fluctuations across the different phases. For both \ac{fd} and \ac{fli} tasks, the model exhibited nearly identical performance to the baseline, even when data from individual phases were missing. In the case of the \ac{fd} task, slight increases in the F1-score were observed for phases A and C ($+0.01\%$), while phase B showed no change compared to the baseline. For the \ac{fli} task, phase A demonstrated a small improvement of \(+0.04\%\), whereas phases B and C exhibited a minimal rise of \(+0.01\%\).

\begin{figure}[htbp]
    \centerline{\includegraphics[width=0.96\linewidth]{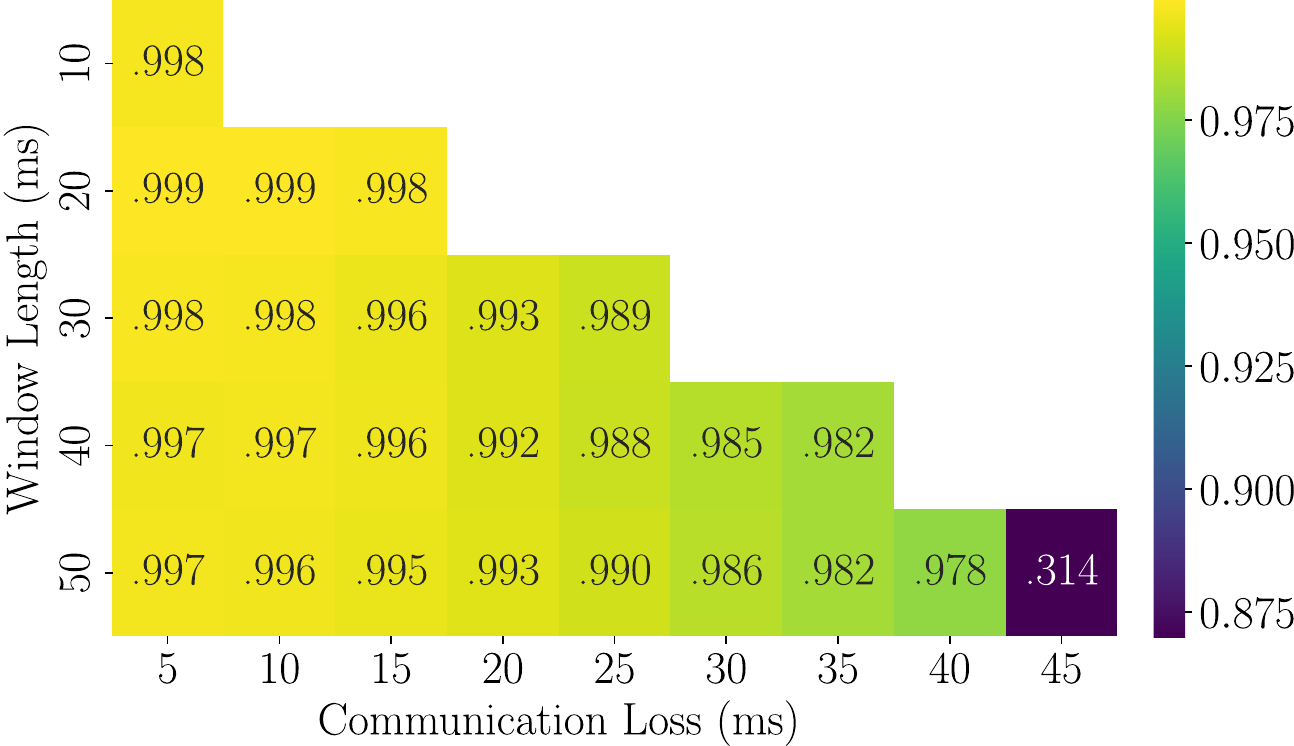}}
    \caption{Fault line identification F1-scores under temporal communication loss. Each gap begins at \(10\%\) of the window duration. Blank cells denote gaps at least as long as the window, which were excluded.}
    \label{fig:heatmap_fault_target_temporal_communication_loss}
\end{figure}

\subsubsection*{Communication Loss}
Short communication losses had a negligible effect on the model's \ac{fd} performance. For outages of up to \(\qty{20}{\milli\second}\), the F1-score remained nearly unchanged, with a maximum drop of only $-0.13\%$. As the duration of the outage increased, the performance gradually decreased, reaching a drop of $-0.90\%$ at \(\qty{40}{\milli\second}\). However, a \qty{45}{\milli\second} communication loss caused a substantial drop in F1-score by $32.8\%$, reducing it to $0.672$.

The \ac{fli} task was more sensitive to temporary communication losses, as shown in Fig.~\ref{fig:heatmap_fault_target_temporal_communication_loss}. Even a brief outage of \(\qty{30}{\milli\second}\) led to a performance decline of $-1.20\%$, reducing the F1-score to \(0.985\). For a \(\qty{45}{\milli\second}\) outage, performance dropped sharply to \(0.314\), corresponding to a decrease of $-68.6\%$. 
The \(\qty{45}{\milli\second}\) gap is evaluated only in a \(\qty{50}{\milli\second}\) window and occupies \(90\%\) of its duration. In that window, a \(\qty{40}{\milli\second}\) gap leaves the final \(\qty{5}{\milli\second}\) visible, while a \(\qty{45}{\milli\second}\) gap removes it as well. The sharp drop occurs with this placement, where only the first \(\qty{5}{\milli\second}\) remain visible.

\section{Discussion}\label{sec:discussion} 
% FD task robost against data sparsity
The random forest retains high \ac{fd} scores under most of the evaluated input restrictions. The spread across window-specific means summarizes variation across the five evaluated durations.
% Missing Voltage or Current Data  
Both tasks retain high F1-scores when voltage measurements are unavailable. Removing current measurements substantially reduces \ac{fli} performance in the evaluated three-phase fault setting.
% Reduced Temporal Resolution  
The random forest maintained similar F1-scores on the evaluated decimated inputs, including at an effective rate of \(\qty{400}{\hertz}\).
% Component Failures  
Component failures had little effect on \ac{fd} in the evaluated scenarios. The Bus~2 condition produced the largest \ac{fli} decrease among the bus-loss scenarios. The historical mapping of those conditions to physical relay channels remains unverified, so the result does not establish which physical bus is most informative.
% Phase-Specific Data Loss  
Phase-specific data losses had minimal impact. This may be due to the symmetrical nature of three-phase faults, reducing the dependency on individual phase measurements for \ac{fd} and \ac{fli}.  
% Communication Loss Impact  
For temporary communication losses, \ac{fd} is less sensitive than \ac{fli} to the evaluated gaps. A \(\qty{5}{\milli\second}\) gap in a \(\qty{10}{\milli\second}\) window had less impact on \ac{fli} than a \(\qty{25}{\milli\second}\) gap in a \(\qty{50}{\milli\second}\) window. Both gaps occupy half of their window, but the comparison changes both the window duration and the gap duration.

% Key Findings and Implications  
Among the evaluated restrictions, \ac{fli} is especially sensitive to missing current measurements, the Bus~2 loss condition, and extended communication gaps. The \ac{fd} scores are less sensitive across most of these conditions.

\section{Conclusion}\label{sec:conclusion}
This study demonstrates the effectiveness of a systematic evaluation strategy for assessing the impact of data sparsity on the performance of ML-based \ac{fd} and \ac{fli}. This approach addresses a critical gap in current research and offers insights that could guide the development of more robust \ac{ml} applications in this field.
Furthermore, it offers comprehensive quantitative findings using an existing ML method, focusing on three-phase short circuits in transmission lines, resulting in a proof of concept for evaluating \ac{ml} models under real-world data constraints.
Our quantitative results show that the \ac{ml} model for \ac{fd} remains robust against sparse and missing data, while the model for \ac{fli} is more sensitive to data sparsity.
The \ac{fd} F1-score remains similar after decimation under the evaluated three-phase fault conditions.
Future research should expand this validation framework to assess how data sparsity affects various \ac{ml} approaches and apply it to different grid topologies.
Additionally, the framework should be extended to include additional fault types (e.g., two-phase, ground faults), load conditions, multiple fault scenarios, and resilience to noisy data.
This would enable more comprehensive, reliable, and real-world applicable validation of \ac{ml} models for power system protection.
%Holi
\section*{Acknowledgment}
This project was funded by the Deutsche Forschungsgemeinschaft (DFG, German Research Foundation) - 535389056.

\bibliographystyle{IEEEtran}
\bibliography{refs}

\end{document}